# Phase Transition for Random Quantified XOR-Formulas

**Nadia Creignou**                                                   CREIGNOU@LIF.UNIV-MRS.FR
*LIF, UMR CNRS 6166*
*Université de la Méditerranée*
*163, avenue de Luminy*
*13 288 Marseille, France*

**Hervé Daudé**                                                      DAUDE@GYPTIS.UNIV-MRS.FR
*LATP, UMR CNRS 6632*
*Université de Provence*
*39, rue Joliot-Curie*
*13 453 Marseille, France*

**Uwe Egly**                                                         UWE@KR.TUWIEN.AC.AT
*Institut für Informationssysteme 184/3*
*Technische Universität Wien*
*Favoritenstraße 9–11*
*A–1040 Wien, Austria*

## Abstract

The QXOR-SAT problem is the quantified version of the satisfiability problem XOR-SAT in which the connective exclusive-or is used instead of the usual or. We study the phase transition associated with random QXOR-SAT instances. We give a description of this phase transition in the case of one alternation of quantifiers, thus performing an advanced practical *and* theoretical study on the phase transition of a quantified problem.

## 1. Introduction

The last decade has seen a growth of interest in the phase transition for Boolean satisfiability (SAT). The clausal version of this problem shows a sharp transition in the sense that, as the density of clauses is increased, formulas abruptly change from being satisfiable to being unsatisfiable at a critical threshold point. Numerous experimental studies have been performed in the investigation of phase transition for different variants of SAT problems, thus giving strong evidence that the location of the transition coincides with such instances that are hard to solve. In the meantime, theoretical studies have been conducted in order to better understand such transitions. Determining the nature of the phase transition (sharp or coarse)[1], locating it, determining a precise scaling window and better understanding the structure of the space of solutions turn out to be very challenging tasks, which have aroused a lot of interest in different communities, namely mathematics, computer science and statistical physics (see e.g., Dubois, Monasson, Selman, & Zecchina, 2001). From a computer science point of view, the success of the SAT problem is due to two features. The problem SAT provides a framework in which problems within the complexity class NP can be adequately expressed, and moreover practically efficient and highly optimized solvers are available.

---

1. Definitions of sharp and coarse phase transitions can be found on p. 21 in (Janson, Luczak, & Rucinski, 2000).

In order to obtain even stronger systems, many researchers have turned to a powerful generalization of Boolean satisfiability, QSAT, where both universal and existential quantifiers over Boolean variables are permitted. The QSAT problem permits the adequate representation and modeling of problems having higher complexity—within the complexity class PSPACE—and coming from various fields of computer science such as knowledge representation, verification and logic. Recently, random instances of these quantified problems have started to attract some attention (see Gent & Walsh, 1999; Chen & Interian, 2005). Models for generating random instances have been developed, and experimental studies have shown that in these models the QSAT property undergoes a phase transition that is qualitatively similar to the one that appears for the ordinary SAT property. As stated by Chen and Interian (2005), the hope is that research on developing competitive solvers for quantified Boolean formulas could benefit from a better understanding of the typical behavior of random instances. Our study follows the pioneering work from Chen and Interian (2005) who have made precise a promising model for generating random instances of the QSAT problem. We use their model and apply it to the satisfiability problem QXOR-SAT that we present below.

The difficulty of identifying transition factors and of performing theoretical explorations of the SAT transition has incited many researchers to turn to a variant of the SAT problem: the e-XOR-SAT problem. This satisfiability problem deals with Boolean formulas in conjunctive normal form with $e$ variables per clause, in which the "usual or" is replaced by the "exclusive or" (we call clauses with "exclusive or" as the only connective XOR-clauses). This problem has contributed to develop or validate techniques, thus revealing typical behaviors of both random instances and their space of solutions for SAT-type problems (see, e.g., Cocco, Dubois, Mandler, & Monasson, 2003; Creignou & Daudé, 2003; Dubois, Boufkhad, & Mandler, 2000; Dubois & Mandler, 2002; Franz, Leone, Ricci-Tersenghi, & Zecchina, 2001). Therefore, in order to understand how phase transitions evolve for satisfiability when introducing quantified variables, it is quite natural to consider this problem.

Although the phase transition of random k-SAT is not yet well understood, generalization to the quantified version has started for some years. The hope is that a generalization of the problem can help in understanding the original one. Another way of gaining insight into a hard problem is to look at some tractable variants. For these reasons in this paper we embark on a theoretical study of the phase transition for the QXOR-SAT problem, which is the quantified version of the XOR-SAT problem. Let us emphasize that this quantified problem is, as the usual XOR-SAT problem, in P, and hence will permit us to provide experiments at a large scale, thus giving useful intuition on the asymptotical behavior of random instances. In order to efficiently solve an instance of the XOR-SAT problem, the clause set is rewritten to a set of equations with coefficients from the finite field $GF(2)$ and Gaussian elimination is performed on the resulting set of equations. Gaussian elimination followed by an examination of the quantifier structure provides an algorithm for the quantified version of the XOR-SAT problem (for details, see Creignou, Khanna, & Sudan, 2001, chap. 6.4).

Following the previous studies conducted by Gent and Walsh (1999) as well as Chen and Interian (2005), we focus on formulas in conjunctive normal form having two quantifier blocks, namely on formulas of the type $\forall X \exists Y \varphi(X, Y)$, where $X$ and $Y$ denote distinct sets of variables, and $\varphi(X, Y)$ is a conjunction of XOR-clauses. Moreover, any variable occurring in $\varphi(X, Y)$ occurs in $X$ or $Y$, i.e., the formula $\forall X \exists Y \varphi(X, Y)$ is closed. The model has several parameters. First we consider (a,e)-QXOR-formulas, which are such that each clause in $\varphi$ has exactly $(a + e)$ variables, $a$ from $X$ and $e$ from $Y$. The (a,e)-QXOR-SAT property is the property for such a formula to be true. The second parameter is a pair $(m, n)$ specifying the number of variables in each quantifier block, i.e., in $X$ and $Y$. The third parameter is $L$, the number of clauses. To sum up, the generated formulas are of the form $\forall X \exists Y \varphi(X, Y)$, where $X$ has $m$ variables, $Y$ has $n$ variables, each clause in $\varphi$ has $a$ variables from $X$ and $e$ from $Y$ and there is a total number of $L$ clauses in $\varphi$. We are interested in the probability that a formula drawn at random uniformly out of this set of formulas is true as $n$ tends to infinity (Section 2). We prove that the nature of the phase transition (coarse or sharp) for (a,e)-QXOR-SAT is governed by the number of existential variables occurring in each clause. For $e = 2$ and any $a \geq 1$, we prove in Section 3 that the (a,2)-QXOR-SAT has a coarse phase transition. Moreover we give an expression of the distribution function of the threshold for (a,2)-QXOR-SAT and we show how it is influenced by the different parameters of the model. For $e \geq 3$, we prove in Section 4 that (a,e)-QXOR-SAT has a sharp phase transition—thus getting the first proof of a sharp threshold for a natural quantified satisfiability problem.

## 2. QXOR, XOR and the Maximal rank Property

In this section, we relate QXOR and XOR to the Maximal rank property. We start with some definitions and notations.

### 2.1 Notation

An *e-XOR-clause*, $C$, is a linear equation over the finite field $GF(2)$ using exactly $e$ distinct variables, $C = ((x_1 \oplus \ldots \oplus x_e) = \varepsilon)$ where $\varepsilon = 0$ or 1.

An *e-XOR-formula*, $\varphi$, is a conjunction of not necessarily distinct $e$-XOR-clauses. A *truth assignment $I$* is a mapping that assigns 0 or 1 to each variable in its domain, it satisfies an XOR-clause $C = ((x_1 \oplus \ldots \oplus x_e) = \varepsilon)$ if and only if $I(C) := \sum_{i=1}^{e} I(x_i) \bmod 2 = \varepsilon$ and it satisfies a formula $\varphi$ if and only if it satisfies every clause in $\varphi$.

We will denote by e-XOR-SAT the property for an $e$-XOR-formula of being satisfiable.

An (a,e)-QXOR-*formula* is a closed quantified formula of the following type

$$\forall X \exists Y \varphi(X, Y),$$

where $X$ and $Y$ denote distinct set of variables, $\varphi(X, Y)$ is an $(a + e)$-XOR-formula such that each clause contains exactly $a$ variables from $X$ and exactly $e$ variables from $Y$. Such a formula is *true* if, for every assignment to the variables $X$, there exists an assignment to the variables $Y$, such that $\varphi$ is true. Observe that, for closed formulas, the notions of truth and satisfiability coincide. For this reason, we will use the two notions synonymously in the following. We denote by (a,e)-QXOR-SAT the property for an (a,e)-QXOR-formula of being true.

Throughout the paper, we reserve $m$ for the number of universal variables (resp. $n$ for the number of existential variables), and $\{x_1, \ldots, x_m\}$ (resp. $\{y_1, \ldots, y_n\}$) denotes the set of such variables. Note that there are

$$N = \binom{m}{a} \cdot \binom{n}{e} \cdot 2 \tag{1}$$

(a,e)-XOR-clauses. We consider random formulas $\forall X \exists Y \varphi(X, Y)$ obtained by choosing uniformly independently and with replacement $L$ clauses from all the possible $N$ (a,e)-XOR-clauses. Using the terminology of Chen and Interian (2005), these formulas correspond to (a,e)-QXOR((m,n),L)-formulas. We are interested in estimating the probability that a randomly chosen (a,e)-QXOR((m,n),L)-formula is true. We denote this probability by $\mathsf{Pr}_{(m,n,L)}$((a,e)-QXOR-SAT), or shortly $\mathsf{Pr}$((a,e)-QXOR-SAT) when no confusion can arise. When restricted to the non-quantified case, e-XOR-SAT, i.e., when a=0, we omit the first component in the subscript, thus discussing $\mathsf{Pr}_{n,L}$(e-XOR-SAT), or shortly $\mathsf{Pr}$(e-XOR-SAT).

We will show that the behavior of the (a,e)-QXOR-SAT property is bounded from above and below by two monotone properties, namely the e-XOR-SAT property and the Maximal rank property. Experiments will suggest that the right parameter in order to study these properties is $c$, the ratio of the number of clauses over the number of existential variables. Moreover, according to the results obtained for e-XOR-SAT by Creignou, Daudé, and Dubois (2003), we know that the transition occurs when $c < 1$. Therefore, in the sequel we will always suppose without loss of generality that $L \leq n$.

## 2.2 Upper and Lower Bounds for the QXOR-SAT Property

Note that a random (a,e)-QXOR((m,n),L)-formula can also be considered as the quantified linear system

$$\forall X \exists Y \quad (AX + EY = C) \tag{2}$$

with coefficient arithmetic in $GF(2)$, where $A$ (respectively $E$) is a matrix chosen uniformly from the set of Boolean $L \times m$ (resp. $L \times n$) matrices with exactly $a$ (respectively $e$) units in each row, and $C$ is a Boolean column vector of dimension $L$ chosen uniformly from the set of all such vectors. Moreover, $A$, $E$ and $C$ are chosen independently.

Observe that the quantified linear system

$$\forall X \exists Y \quad (AX + EY = C)$$

is consistent if and only if

$$\forall X \quad (C - AX) \in \mathsf{Im}(E),$$

where $\mathsf{Im}(E)$ represents the image of the linear application whose matrix representation is $E$, that is $\mathsf{Im}(E) = \{EY / \ Y \in \{0,1\}^n\}$. Hence the quantified linear system is consistent if and only if $C \in \mathsf{Im}(E)$ and $\mathsf{Im}(A) \subseteq \mathsf{Im}(E)$. Therefore, we get:

$$\begin{aligned} \mathsf{Pr}((\mathsf{a,e})\text{-QXOR-SAT}) &= \mathsf{Pr}(\forall X \exists Y \quad (AX + EY = C) \text{ is consistent}) \\ &= \mathsf{Pr}(C \in \mathsf{Im}(E) \text{ and } \mathsf{Im}(A) \subseteq \mathsf{Im}(E)). \end{aligned}$$

Thus, on the one hand

$$\mathsf{Pr}((\mathsf{a,e})\text{-QXOR-SAT}) \leq \mathsf{Pr}(C \in \mathsf{Im}(E)) = \mathsf{Pr}(\mathsf{e}\text{-XOR-SAT})$$

holds, and on the other hand, the inequality

$$\mathsf{Pr}((\mathsf{a,e})\text{-QXOR-SAT}) \geq \mathsf{Pr}(\mathsf{Im}(E) = \{0,1\}^L)$$

holds. Therefore, if $\mathsf{Pr}_{n,L}(\mathsf{e\text{-}Max\text{-}rank})$ denotes the probability that a random matrix from the set of $L \times n$ Boolean matrices with $e$ units per row is of maximal rank, then for every $m \geq a$, all $n$ and all $L \leq n$, we get the following inequalities:

$$\mathsf{Pr}_{n,L}(\mathsf{e\text{-}Max\text{-}rank}) \leq \mathsf{Pr}_{(m,n,L)}((\mathsf{a,e})\text{-QXOR-SAT}) \leq \mathsf{Pr}_{n,L}(\mathsf{e\text{-}XOR\text{-}SAT}). \qquad (3)$$

A natural question at this stage is to estimate the probability that a random matrix is of maximal rank. In the following we will provide some experiments and theoretical results comparing the behavior of the three properties, Maximal rank, QXOR-SAT and XOR-SAT, thus making precise the behavior of the (a,e)-QXOR-SAT property according to the value of $e$.

## 3. The Case $e = 2$

In this section, we restrict our attention to the case where all problems have two existential variables in each clause (and the number of all variables is allowed to vary).

### 3.1 Experimental Results

In order to illustrate the inequalities (3) and to compare empirically the three properties, we discuss experiments that we have performed. In the experiments, all formulas are closed.

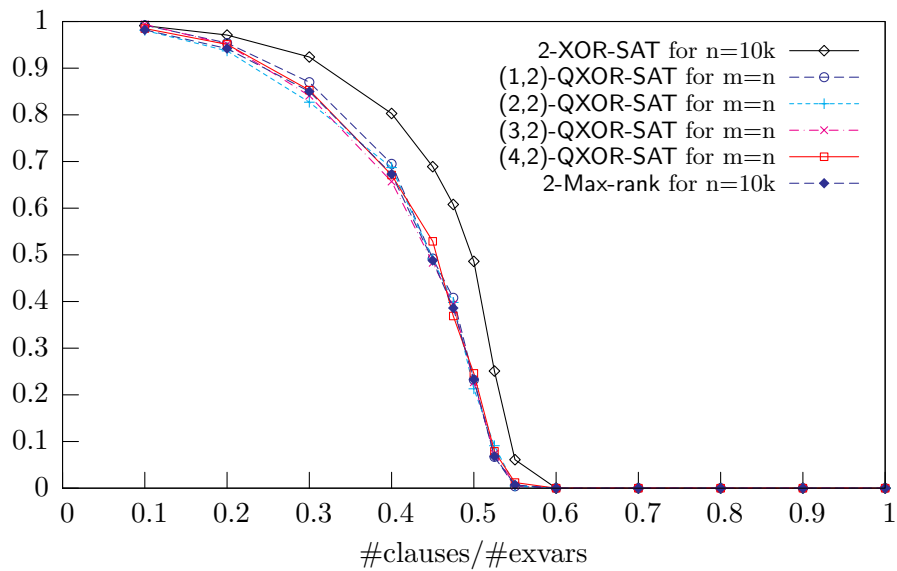

Figure 1: The curves for (a,2)-QXOR-SAT, $m = n = 10\ 000$ and $a$ varying.

In all cases, the experiments have been conducted according to the same scheme. Let us describe it in detail for Figure 1. One experiment consisted in generating at random (in drawing uniformly and independently) (1,2)-QXOR-formulas over 10 000 existential variables and 10 000 universal variables with a ratio "number of clauses/number of existential variables" varying from 0.1 to 1 in steps of 0.05 or 0.1. For each of the chosen values of ratio, a sample of 1000 formulas were studied using the algorithm described in the work of Creignou et al. (2001, chap. 6.4), thus deciding their truth (or satisfiability) as quantified formulas. The proportion of true instances for each considered value of ratio has been plotted in Figure 1. The same has been done for the other (a,2)-QXOR-SAT properties. Hence, the different curves are independent from each other. For the 2-XOR-SAT experiment, we used the same selection procedure over 10 000 existential variables. Again, Gaussian elimination together with an examination of the quantifier structure were used to determine the logical "status" (true or false) of every formula. Additionally, it has been computed whether the matrix $E$ has full rank or not. The curve 2-Max-rank shows the proportion of systems with full rank and it corresponds to the 2-XOR-SAT curve in the same figure. A close look at Figure 1 reveals that some points from the (a,2)-QXOR-SAT curves are slightly *below* the (theoretical) lower bound given by the curve for 2-Max-rank. The reason for this phenomenon is the independence of all the satisfiability curves from each other and the "noise" induced by the finite sampling of problems. If we had chosen corresponding problems with exactly the same existential part (and only the universal part varies), then we would have got all satisfiability curves above the curve for 2-Max-rank.

The experimental results shown in Figure 1 suggest first that the two bounding properties, namely 2-Max-rank and 2-XOR-SAT are distinguishable, second that, when $m = n$, the (a,2)-QXOR-SAT property coincides asymptotically with the 2-Max-rank property independently of $a \geq 1$, the number of universal variables per clause.

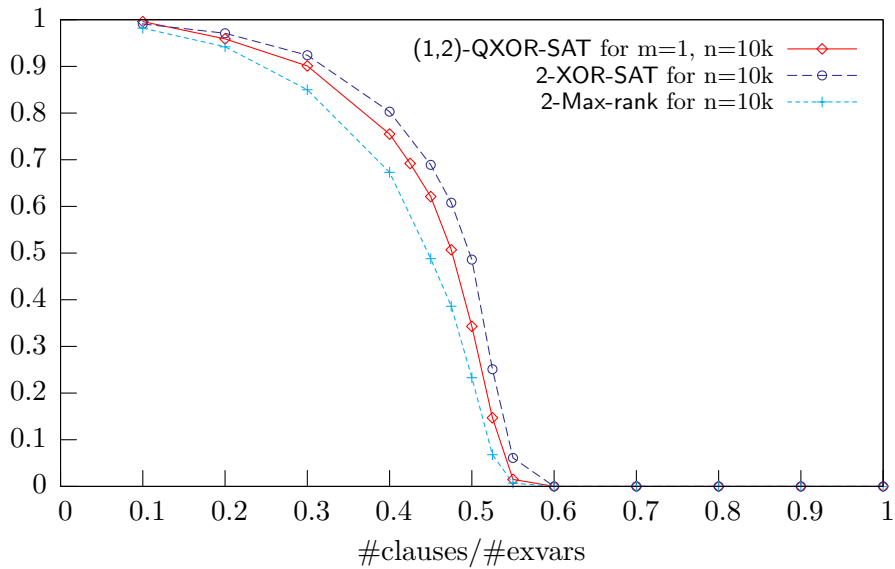

Figure 2: (1,2)-QXOR-SAT when $m = 1$ and $n = 10\,000$ compared to 2-XOR-SAT and 2-Max-rank.

At another scale for $m$, for instance when $m$ is constant, the experimental results reported in Figure 2 suggest that the (a,2)-QXOR-SAT property is in between the two properties 2-Max-rank and 2-XOR-SAT.

In the following, we will validate the conjectures suggested by these experiments and prove that the property (a,2)-QXOR-SAT coincides asymptotically with the 2-Max-rank property as soon as $m$ is tending to infinity with $n$, and that it is in between the 2-Max-rank property and the 2-XOR property when $m$ is a fixed constant. In particular, we will make clear the connection between random (a,2)-QXOR-formulas and random labelled graphs.

### 3.2 Bad Cycles and the (a,2)-QXOR-SAT Property

We are interested in the satisfiability of quantified systems of the form

$$s = (\forall X \exists Y \ AX + EY = C),$$

where $E$ (respectively $A$) is a matrix chosen uniformly in the set of Boolean $L \times n$ (respectively $L \times m$) matrices with exactly 2 (respectively $a$) units in each row, and $C$ is a random column vector of dimension $L$, in which 0 and 1 occur with the same probability. The satisfiability of such a system is strongly related to the existence of cycles in graphs. Indeed, we construct a graph $G_a(s)$ with $n$ vertices and $L$ weighted edges. For each existential variable $y_i$ we have a vertex in $G_a(s)$. For each equation $y_i \oplus y_j = x_{i_1} \oplus \ldots \oplus x_{i_a} \oplus \varepsilon$, we add the edge $\{y_i, y_j\}$ to $G_a(s)$ with the label $x_{i_1} \oplus \ldots \oplus x_{i_a} \oplus \varepsilon$. A cycle is given by a sequence of vertices $(y_{i_1}, \ldots, y_{i_s})$ such that for $1 \leq j \leq s-1$, $\{y_{i_j}, y_{i_{j+1}}\}$ is an edge in the graph, and so is $\{y_{i_s}, y_{i_1}\}$. The cycle is said to be *elementary* if all the vertices in the sequence are distinct. The *weight of a cycle* is the sum modulo 2 of the labels of its edges.

**Example 3.1** *Let $X = \{x_1, x_2, x_3\}$ and let $Y = \{y_1, \ldots, y_7\}$. The formula $\forall X \exists Y \ \varphi(X, Y)$ with $\varphi(X, Y)$ being a conjunction of the following equations*

| | | | | | | |
|---|---|---|---|---|---|---|
| $y_1 \oplus y_2$ | $=$ | $x_1$ | | $y_1 \oplus y_7$ | $=$ | $x_2$ |
| $y_2 \oplus y_3$ | $=$ | $x_3$ | | $y_2 \oplus y_6$ | $=$ | $x_2 \oplus 1$ |
| $y_3 \oplus y_4$ | $=$ | $x_2 \oplus 1$ | | $y_3 \oplus y_5$ | $=$ | $x_3$ |
| $y_4 \oplus y_5$ | $=$ | $x_3 \oplus 1$ | | $y_6 \oplus y_7$ | $=$ | $x_1 \oplus 1$ |

*can be represented by the graph in Figure 3.*

In the following, we call a cycle *bad* when it has a nonzero weight, and *good* otherwise.

**Example 3.2** *In the graph associated with the formula described in the previous example, there is a good cycle, $(y_1, y_2, y_6, y_7)$, whose weight is 0, and a bad one, $(y_3, y_4, y_5)$, whose weight is $x_2$. For the latter cycle, the corresponding equations are $y_3 \oplus y_4 = x_2 \oplus 1$, $y_3 \oplus y_5 = x_3$, and $y_4 \oplus y_5 = x_3 \oplus 1$. Adding these three equations yields the equation $0 = x_2$ which cannot be satisfied because $x_2 \in X$ is universally quantified.*

For systems containing only existential variables, i.e., $a = 0$, it has been observed by Creignou and Daudé (2003) that a 2-XOR-formula is satisfiable if and only if the graph $G_0(s)$ has no bad cycle, that is :

$$\Pr(\text{2-XOR-SAT}) = \Pr(G_0(s) \text{ has no bad cycle}). \qquad (4)$$

Using similar arguments, we get the following proposition.

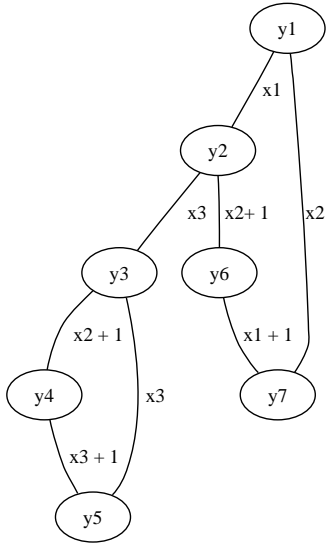

Figure 3: The graph $G_a(s)$ from Example 3.1 (addition is performed mod2).

**Proposition 3.3** *The system $s$ is satisfiable if and only if $G_a(s)$ does not contain any elementary bad cycle, i.e.,*

$$\mathsf{Pr}((\mathsf{a},2)\text{-}\mathsf{QXOR\text{-}SAT}) = \mathsf{Pr}(G_a(s) \text{ has no bad cycle}).$$

**Proof:** Suppose we have an elementary cycle with nonzero weight in $G_a(s)$. Clearly, to such a cycle corresponds a subsystem in $s$, for which there exists an assignment to $X$ such that no assignment to $Y$ will satisfy it (see Example 3.2 for an illustration). Hence, $s$ is unsatisfiable.

Conversely, suppose there is no (elementary) cycle with nonzero weight in $G_a(s)$. Take an arbitrary truth assignment $I$ for the (universal) variables in $X$ and apply it to $G_a(s)$. Since $I(x) \in \{0,1\}$ for all $x \in X$, the weight at each edge can be reduced to a constant from $\{0,1\}$ by addition modulo 2. Moreover, since each cycle in $G_a(s)$ has zero weight, the corresponding cycle in the "reduced" version, $G_a'(s)$, of $G_a(s)$ has also zero weight. The graph $G_a'(s)$ corresponds to a system with existential quantifiers only.

In order to obtain a satisfying truth assignment for the existential variables, it suffices to apply the following procedure to each connected component of $G_a'(s)$. Consider an arbitrary root vertex $y$ and assign an arbitrary truth value $v$ to it. We obtain a truth value for each vertex in $G_a'(s)$ by performing a depth-first search starting from $y$. During the search, if there is an edge $(y', y'')$ labelled with $\varepsilon$ and $y''$ has no truth value yet, then we set the value for $y''$ to the value of $y' \oplus \varepsilon$. The assignment obtained in this way satisfies all the equations since $G_a'(s)$ does not contain any cycle with nonzero weight. ∎

**Remark:** Observe that

$$\mathsf{Pr}(2\text{-}\mathsf{Max\text{-}rank}) = \mathsf{Pr}(G_a(s) \text{ has no cycle}) \tag{5}$$

holds.

### 3.3 The Distribution Functions for 2-XOR-SAT and 2-Max-rank

In this section, we will give the exact asymptotical value of the bounds obtained in (3) in terms of the order parameter $c$, where $c \cdot n$ is the number of clauses. For this we will use well-known results from random graph theory.

Let us recall that we consider the classical probabilistic model where each clause/edge is chosen uniformly and independently among the

$$N = \binom{m}{a} \cdot \binom{n}{e} \cdot 2$$

possible ones. According to Proposition 1.13 in (Janson et al., 2000), if we choose $L = c \cdot n$ clauses, then this model is asymptotically equivalent to the one where each clause is drawn independently with probability $p$, where

$$p = \frac{c \cdot n}{\binom{m}{a} \cdot \binom{n}{2} \cdot 2} \left(1 + O(n^{-1/2})\right)$$

holds. So, in the following, we will work with the random labelled graphs $G_a(s)$ associated with quantified systems $s$, with labelled edge probability:

$$p = \frac{c}{n \cdot \binom{m}{a}}.$$

The corresponding probability is usually denoted by $\mu_p$. However, for simplicity we will keep the notation $\mathsf{Pr}$.

In the light of Proposition 3.3 and of (5), it appears that we have to study

$$\mathsf{Pr}(G_a(s) \text{ has no (bad) cycle}).$$

The asymptotic behavior of the number of cycles in random graphs has been first investigated by Erdős and Rényi (1960), and made precise by Janson (1987) and Takács (1988). This number converges in distribution to a Poisson law of parameter $\lambda$, where $\lambda$ is the limit of the average number of cycles as $n$, the number of vertices, tends to infinity.

This result can be easily extended to our model of labelled graphs. In particular,

$$\mathsf{Pr}(G_a(s) \text{ has no (bad) cycle}) \longrightarrow \exp(-\lambda),$$

where $\lambda$ is the limit of the the average number of (bad) cycles. A challenging task is now to get a simple expression of lambda.

Let $Y$ be the random variable that counts the number of cycles. Let $\mathcal{C}$ be the set of all possible cycles. For any cycle $c$, we introduce the random variable $X_c$ such that $X_c(G_a(s)) = 1$ holds, if and only if $G_a(s)$ contains the cycle $c$. The average number of cycles is

$$E(Y) = E(\sum_{c \in \mathcal{C}} X_c) = \sum_{c \in \mathcal{C}} E(X_c).$$

Since every cycle $c$ of length $l$ has expectation $E(X_c) = p^l$ and since the number of cycles of length $l$ is $\dfrac{n(n-1)\dots(n-l+1)}{2l} \cdot \binom{m}{a}^l \cdot 2^l$, we get that

$$E(Y) = \sum_{l \geq 2} \frac{n(n-1)\dots(n-l+1)}{2l} \cdot \binom{m}{a}^l \cdot 2^l \cdot p^l$$

holds. Thus,

$$\lim_{n \to +\infty} E(Y) = \sum_{l \geq 2} \frac{(2c)^l}{2l} = -\frac{1}{2}\ln(1-2c) - c$$

also holds. From (5), we obtain for every $0 < c < \frac{1}{2}$ that

$$\lim_{n \to +\infty} \mathsf{Pr}_{(n,cn)}(\mathsf{2\text{-}Max\text{-}rank}) = \exp(\frac{1}{2}\ln(1-2c) + c)$$

holds, and finally

$$\lim_{n \to +\infty} \mathsf{Pr}_{(n,cn)}(\mathsf{2\text{-}Max\text{-}rank}) = H_\infty(c) \tag{6}$$

is established, where

$$H_\infty(c) = \begin{cases} \exp(c) \cdot (1-2c)^{1/2} & \text{for } 0 \leq c \leq \frac{1}{2}, \\ 0 & \text{otherwise.} \end{cases}$$

The experimental results shown in Figure 4 illustrate this asymptotical behavior.

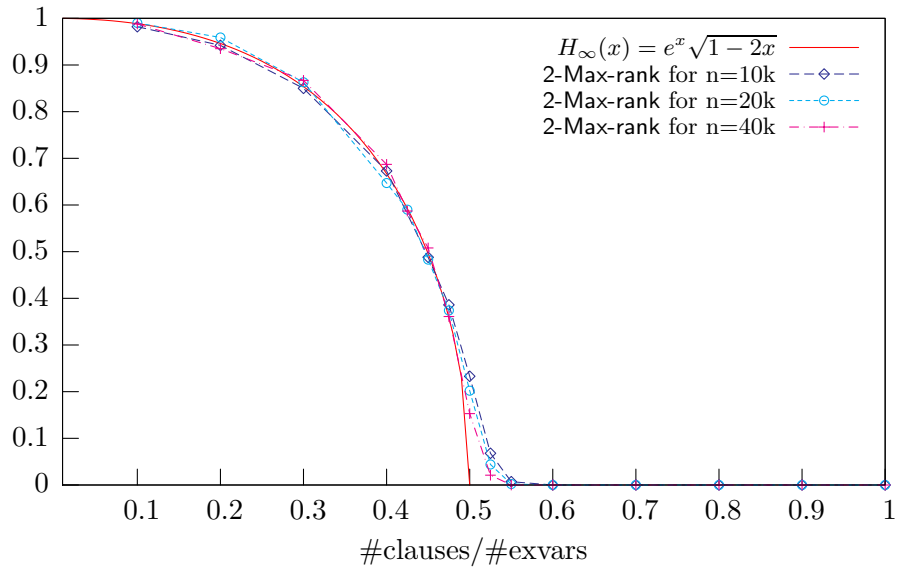

Figure 4: The curves for the 2-Max-rank property.

According to (4)

$$\lim_{n \to +\infty} \mathsf{Pr}_{(n,cn)}(\mathsf{2\text{-}XOR\text{-}SAT}) = \exp(-\lambda_0)$$

holds, where $\lambda_0$ denotes the limit of the average number of bad cycles. In this particular case, the weight of a clause is either 0 or 1, which means that half of the cycles are bad. Thus,

$$\lambda_0 = \lim_{n \to +\infty} \sum_{l \geq 2} \frac{n(n-1)\dots(n-l+1)}{2l} \cdot \binom{m}{a}^l \cdot 2^{l-1} \cdot p^l = -\frac{1}{4}\ln(1-2c) - \frac{c}{2}.$$

Therefore, for every $c \geq 0$, we get

$$\lim_{n \to +\infty} \mathsf{Pr}_{(n,cn)}(\text{2-XOR-SAT}) = \exp(\frac{1}{4}\ln(1-2c) + \frac{c}{2}),$$

and finally

$$\lim_{n \to +\infty} \mathsf{Pr}_{(n,cn)}(\text{2-XOR-SAT}) = H_0(c) \tag{7}$$

is established, where

$$H_0(c) = \begin{cases} \exp(c/2) \cdot (1-2c)^{1/4} & \text{for } 0 \leq c \leq \frac{1}{2}, \\ 0 & \text{otherwise.} \end{cases}$$

This is illustrated by Figure 5.

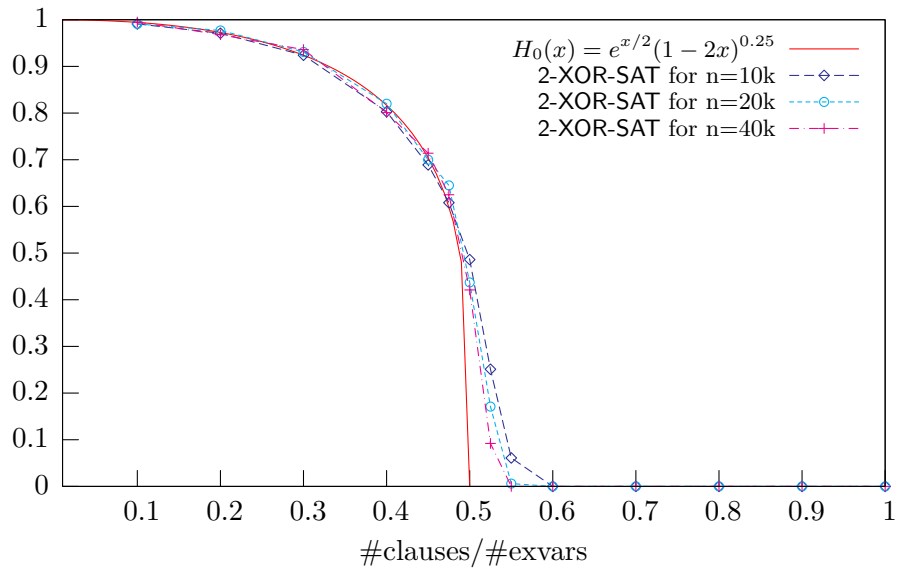

Figure 5: The curves for the 2-XOR-SAT property.

### 3.4 The Distribution Function for (a,2)-QXOR-SAT

The results obtained in the previous section, (6) and (7), together with the inequalities (3) are sufficient to conclude that the (a,2)-QXOR-SAT property has a coarse phase transition at the scale $L = c \cdot n$ and that its distribution function is in between the functions $H_0$ and $H_\infty$ described above. More precisely we get the following theorem.

**Theorem 3.4** *For any integer $a \geq 1$ and every $c \geq 0$, let us consider (a,2)-QXOR((m,n),cn)-formulas consisting in the conjunction of $c \cdot n$ XOR-clauses, where each clause contains $a$ variables from a set of $m$ universal variables, and $e$ variables from a set of $n$ existential variables. Then, when $n$ tends to infinity, the (a,2)-QXOR-SAT property has a coarse phase* *transition whose asymptotical distribution function can be expressed as a function depending on $m$. More precisely, for all $c \geq 0$, every $a \geq 1$ and $m = a$*

$$\mathsf{Pr}_{(a,n,cn)}(\mathsf{(a,2)\text{-}QXOR\text{-}SAT}) \longrightarrow_{n \to +\infty} H(c)$$

*holds, where*

$$H(c) = \begin{cases} \exp(c)(1-2c)^{1/2}(1-4c^2)^{-1/8} & \text{for } 0 \leq c \leq \frac{1}{2}, \\ 0 & \text{otherwise.} \end{cases}$$

*If $m$ is a function of $n$ tending to infinity with $n$, then, for every $a \geq 1$,*

$$\mathsf{Pr}_{(m,n,cn)}(\mathsf{(a,2)\text{-}QXOR\text{-}SAT}) \longrightarrow_{n \to +\infty} H_\infty(c)$$

*holds, where*

$$H_\infty(c) = \begin{cases} \exp(c) \cdot (1-2c)^{1/2} & \text{for } 0 \leq c \leq \frac{1}{2}, \\ 0 & \text{otherwise.} \end{cases}$$

*Moreover, for every fixed $m \geq a \geq 1$, there exists a distribution function $H_m$ such that*

$$\mathsf{Pr}_{(m,n,cn)}(\mathsf{(a,2)\text{-}QXOR\text{-}SAT}) \longrightarrow_{n \to +\infty} H_m(c),$$

*with $H_m$ satisfying*

$$H_\infty < H_m < H_0,$$

*where*

$$H_0(c) = \begin{cases} \exp(c/2)(1-2c)^{1/4} & \text{for } 0 \leq c \leq \frac{1}{2}, \\ 0 & \text{otherwise.} \end{cases}$$

**Proof:** The proof is based on Proposition 3.3 and, as discussed in the previous section, on an estimation of the number of bad cycles in the labelled graphs associated with random formulas.

Let $\lambda_{m,a}$ be the limit of the average number of bad cycles. We will give a closed expression of $\lambda_{m,a}$. Observe that each label on the edges of the graph associated with a $\mathsf{(a,2)\text{-}QXOR\text{-}}((\mathsf{m,n}),\mathsf{cn})$-formula is formed with a constant, 0 or 1, and a "variable-label" made of $a$ universal variables. There are exactly $\binom{m}{a}$ such variable-labels, which are numbered from 1 to $\binom{m}{a}$. One can decide whether a cycle is good or bad according to the number of 1 (even or odd) and the number of occurrences of each variable-label. Therefore it is quite natural to associate to every cycle its length $l$, the sequence $(N_1, \dots, N_{\binom{m}{a}})$ of the numbers of occurrences of each variable-label, and the parity $\varepsilon = 0$ or 1 of the number of occurrences of the constant 1. The limit of the average number of cycles whose parameter $(l, (N_1, \dots, N_{\binom{m}{a}}), \varepsilon)$ is fixed is

$$\frac{c^l 2^{l-1}}{2l} \cdot \frac{\binom{l}{N_1, \dots, N_{\binom{m}{a}}}}{\binom{m}{a}^l}.$$

Moreover, from such a parameter $(l, (N_1, \dots, N_{\binom{m}{a}}), \varepsilon)$, one can decide whether a cycle is bad or not.

For a better readability, let us focus on the case $a = 1$. In this particular case, the label of an edge is of the form $x_i \oplus \varepsilon$, where $\varepsilon = 0$ or 1 and $1 \leq i \leq m$. On the one hand, all cycles of odd length are bad (for in the weight of such a cycle at least one of the coefficients of the $x_i$'s will be nonzero). On the other hand, there are two kinds of cycles of even length that are bad. The ones in which the constant 1 appears an odd number of times, and the ones in which one of the universal variables appears an odd number of times. Since, for $m \geq 1$, we have

$$m^l = \sum_{\forall i \ N_i \equiv 0(2)} \binom{l}{N_1, N_2, \ldots, N_m} + \sum_{\exists i \ N_i \equiv 1(2)} \binom{l}{N_1, N_2, \ldots, N_m}$$

we get

$$\lambda_{m,1} = \sum_{u \geq 1} \frac{(2c)^{2u+1}}{2(2u+1)} + \frac{1}{2} \sum_{v \geq 1} \frac{(2c)^{2v}}{(2(2v))} + \frac{1}{2} \sum_{v \geq 1} \frac{(2c)^{2v}}{(2(2v))} \cdot \frac{\sum_{\exists i \ N_i \equiv 1(2)} \binom{l}{N_1, N_2, \ldots, N_m}}{m^l}. \quad (8)$$

Standard combinatorial computations show that, for even $l$, the equation

$$\frac{\sum_{\exists i \ N_i \equiv 1(2)} \binom{l}{N_1, N_2, \ldots, N_m}}{m^l} = 1 - \sum_{k=0}^{m-1} \binom{m-1}{k} \cdot \frac{(m-2k)^l}{2^{m-1} \cdot m^l}$$

holds. Therefore, we rewrite (8) and obtain

$$\lambda_{m,1} = \sum_{u \geq 1} \frac{(2c)^{2u+1}}{2(2u+1)} + \sum_{v \geq 1} \frac{(2c)^{2v}}{2(2(2v))} + \sum_{v \geq 1} \frac{(2c)^{2v}}{2(2(2v))} \cdot \left( 1 - \sum_{k=0}^{m-1} \binom{m-1}{k} \cdot \frac{(m-2k)^l}{2^{m-1} \cdot m^l} \right).$$

First, observe that $\lambda_{m,1}$ is a function of $c$, thus we deduce the last part of the theorem:

$$\lim_{n \to +\infty} \mathsf{Pr}_{(m,n,cn)}((1,2)\text{-QXOR-SAT}) = \exp(-\lambda_{m,1}) = H_m(c).$$

Second, from the above expression of $\lambda_{m,1}$ and using the following inequality

$$1 - \sum_{k=0}^{m-1} \binom{m-1}{k} \cdot \frac{(m-2k)^l}{2^{m-1} \cdot m^l} \geq 1 - \frac{4}{m},$$

we get that

$$\lim_{m \to +\infty} (\lambda_{m,1}) = \sum_{l \geq 2} \frac{(2c)^l}{2l} = -\frac{1}{2} \ln(1 - 2c) - c$$

holds, which proves the second statement of the theorem.

In addition, when $m = a = 1$, we get the equation

$$\lambda_{1,1} = \sum_{u \geq 1} \frac{(2c)^{2u+1}}{2(2u+1)} + \sum_{v \geq 1} \frac{(2c)^{2v}}{2(2(2v))} = -\frac{1}{2} \ln(1 - 2c) + \frac{1}{8} \ln(1 - 4c^2) - c.$$

Thus, we have established

$$\lim_{n \to +\infty} \mathsf{Pr}_{(1,n,cn)}((\text{1,2})\text{-QXOR-SAT}) = \exp(-\lambda_{1,1}) = H(c),$$

where

$$H(c) = \begin{cases} \exp(c)(1-2c)^{1/2}(1-4c^2)^{-1/8} & \text{for } 0 \le c \le \frac{1}{2}, \\ 0 & \text{otherwise.} \end{cases}$$

This result is illustrated by Figure 6, while Figure 7 shows the comparative behavior of the three distribution functions $H_0$, $H$ and $H_\infty$.

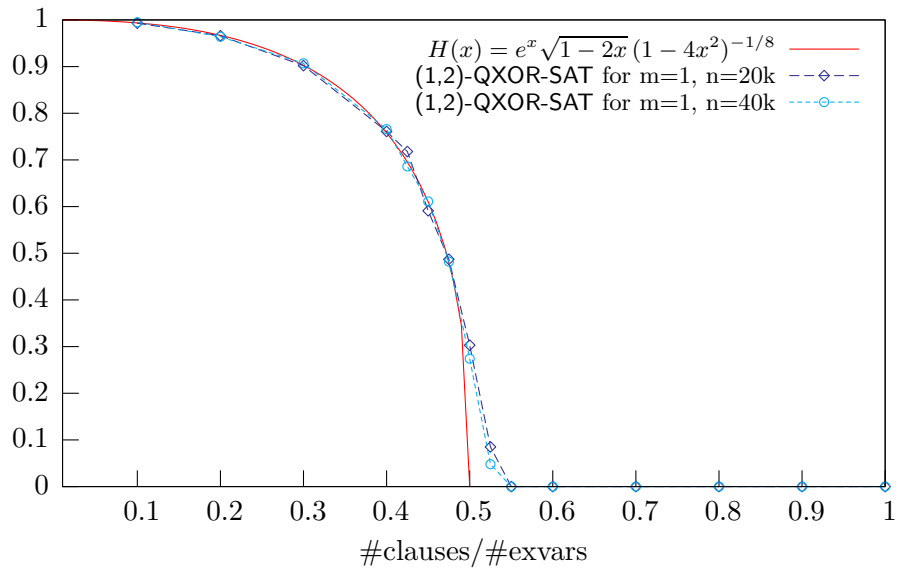

Figure 6: The curves for the (1,2)-QXOR-SAT property with $m = 1$.

## 4. The Case $e \ge 3$

It can be observed from the experimental results shown in Figure 8 that, contrary to what has been observed in the previous section, the three smooth lines connecting the consecutive points and corresponding to the transition of the three properties 3-Max-rank, (a,3)-QXOR-SAT and 3-XOR-SAT are difficult to distinguish. Moreover, when $n$ increases (see Figure 9), the curves straighten and come closer one to each other, showing thus strong empirical evidence that the transitions of the three properties coincide asymptotically, with a sharp phase transition at the critical value $c_3 \approx 0.918$ (which is the critical ratio for 3-XOR-SAT, see Dubois & Mandler, 2002). We will show that, for $e \ge 3$, the introduction of universal variables in XOR-formulas does not influence the sharp transition.

**Theorem 4.1** *For every $e \ge 3$ and any integer $a$, the* (a,e)-QXOR-SAT *property has a sharp threshold which coincides with the one of the* e-XOR-SAT *property.*

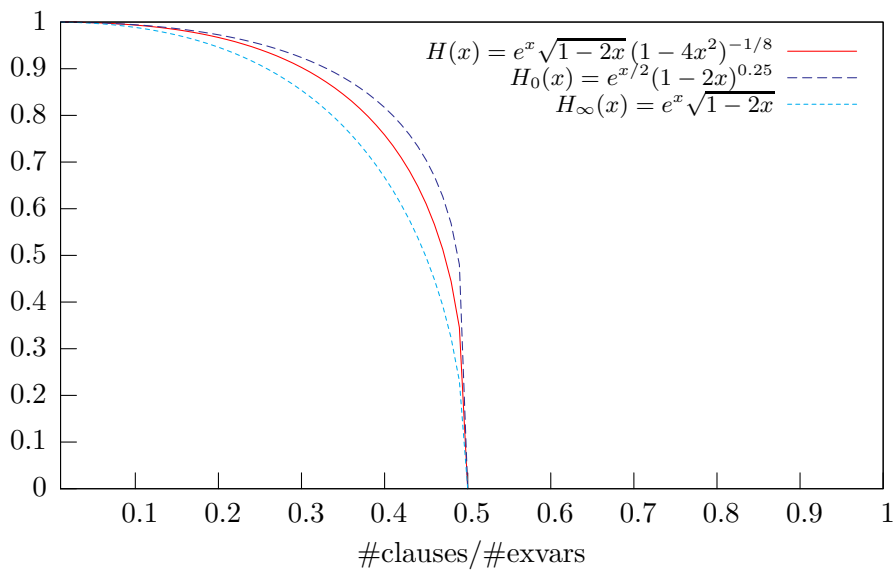

Figure 7: The distribution functions $H_0$, $H$ and $H_\infty$.

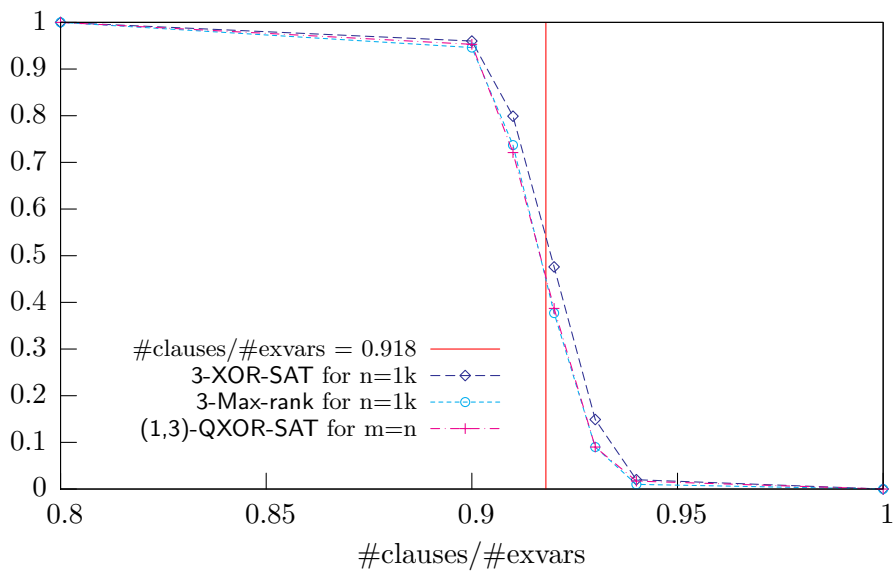

Figure 8: The curves for 3-XOR-SAT, 3-Max-rank and (a,3)-QXOR-SAT for n=1000.

**Proof:** Let us recall that

$$
\begin{aligned}
\mathsf{Pr}_{n,L}(\text{e-XOR-SAT}) &= \mathsf{Pr}(\exists Y \quad (EY = C) \text{ is consistent}) \\
&= \mathsf{Pr}(C \in \mathsf{Im}(E)) \\
&= \sum_{V \subseteq \{0,1\}^L} \mathsf{Pr}(C \in V \text{ and } \mathsf{Im}(E) = V),
\end{aligned}
$$

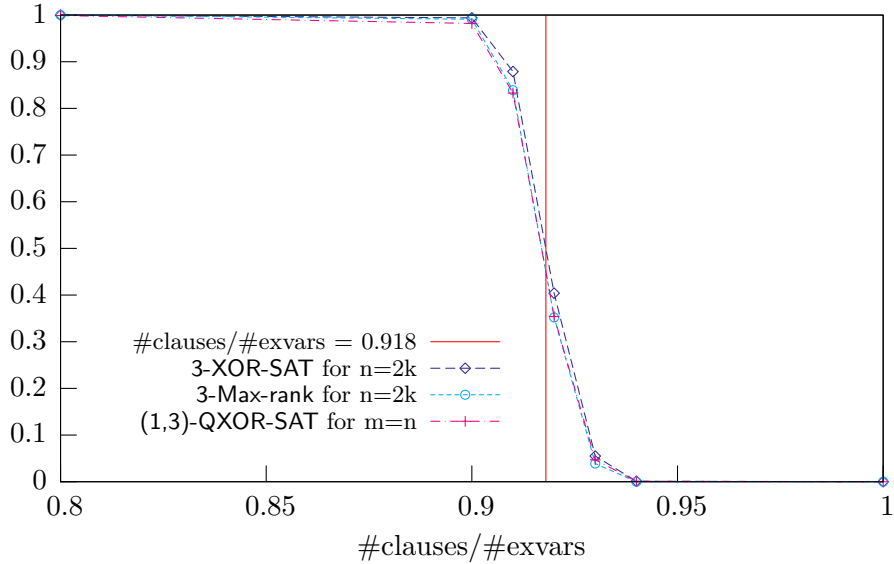

Figure 9: The curves for 3-XOR-SAT, 3-Max-rank and (a,3)-QXOR-SAT for n=2000.

since $E$ and $C$ are chosen independently. Therefore, if $P_{(r)}$ denotes the probability that a random matrix from the set of $L \times n$ Boolean matrices with $e$ units per row is of rank $r$, then

$$\mathsf{Pr}_{n,L}(\mathsf{e\text{-}XOR\text{-}SAT}) = \sum_{r=0}^{L} 2^{r-L} P_{(r)} \leq P_{(L)} + \frac{1}{2}(P_{(L-1)} + \dots P_{(0)}).$$

Now observe that $P_{(0)} + \dots + P_{(L)} = 1$, thus $(P_{(L-1)} + \dots P_{(0)}) = 1 - P_{(L)}$, and hence we get

$$\mathsf{Pr}_{n,L}(\mathsf{e\text{-}XOR\text{-}SAT}) \leq \frac{1 + P_{(L)}}{2}.$$

Therefore, according to (3), we have

$$2 \cdot \mathsf{Pr}_{n,L}(\mathsf{e\text{-}XOR\text{-}SAT}) - 1 \leq \mathsf{Pr}_{n,L}((\mathsf{a,e})\text{-}\mathsf{QXOR\text{-}SAT}) \leq \mathsf{Pr}_{n,L}(\mathsf{e\text{-}XOR\text{-}SAT}).$$

Since we know that the property e-XOR-SAT exhibits a sharp threshold when $L$ is $\Theta(n)$ (Creignou & Daudé, 2003), this shows that (a,e)-QXOR-SAT also does. The same holds for the property e-Max-rank (since $\mathsf{Pr}_{n,L}(\mathsf{e\text{-}Max\text{-}rank}) = P_{(L)}$). In particular, for $e = 3$, we have shown that (a,e)-QXOR-SAT as well as 3-Max-rank have a sharp threshold with a critical value $c_3 \approx 0.918$, which is the critical ratio for 3-XOR-SAT (Dubois & Mandler, 2002). ∎

## 5. Conclusion

We have (experimentally and theoretically) analyzed the phase transition for the quantified problems (a,e)-QXOR-SAT. Our analysis has been conducted at the same level of sophistication as the one made for the e-XOR-SAT problem, thus showing that the model proposed by Chen and Interian (2005) is mathematically tangible and provides the good parameters in order to perform a mathematical analysis of the phase transition for quantified problems.

On the one hand, as observed for QSAT (Gent & Walsh, 1999; Chen & Interian, 2005), we have proved that the nature of the transition is not influenced by the introduction of universal variables. On the other hand, in contrast with QSAT, we have proved that the location of the phase transition—the critical ratio—is the same for the two properties XOR-SAT and QXOR-SAT, and that the difference of behavior between these two properties occurs at the level of the distribution function.

## Acknowledgments

This work has been supported by EGIDE 10632SE and ÖAD Amadée 2/2006.

## References

Chen, H., & Interian, Y. (2005). A model for generating random quantified boolean formulas. In *Proceedings of the 19th International Joint Conference on Artificial Intelligence (IJCAI'2005)*, pp. 66–71.

Cocco, S., Dubois, O., Mandler, J., & Monasson, R. (2003). Rigorous decimation-based construction of ground pure states for spin glass models on random lattices. *Physical Review Letters*, *90*, 472051–472054.

Creignou, N., & Daudé, H. (2003). Coarse and sharp thresholds for random *k*-XOR-CNF. *Informatique théorique et applications/Theoretical Informatics and Applications*, *37*(2), 127–147.

Creignou, N., Daudé, H., & Dubois, O. (2003). Approximating the satisfiability threshold for random *k*-XOR-CNF formulas. *Combinatorics, Probability and Computing*, *12*(2), 113–126.

Creignou, N., Khanna, S., & Sudan, M. (2001). *Complexity classifications of Boolean constraint satisfaction problems*. SIAM Monographs On Discrete Mathematics And Applications. SIAM, Philadelphia, PA, USA.

Dubois, O., Boufkhad, Y., & Mandler, J. (2000). Typical random 3-SAT formulae and the satisfiability threshold. In *Proceedings of the 11th ACM-SIAM Symposium on Discrete Algorithms (SODA'2000)*, pp. 124–126.

Dubois, O., & Mandler, J. (2002). The 3-XOR-SAT threshold. In *Proceedings of the 43th Annual IEEE Symposium on Foundations of Computer Science (FOCS 2002)*, pp. 769–778.

Dubois, O., Monasson, R., Selman, B., & Zecchina, R. (2001). Editorial. *Theoretical Computer Science*, *265*(1–2).

Erdős, P., & Rényi, A. (1960). On the evolution of random graphs. In *Publ. Math. Inst. Hungar. Acad. Sci.*, Vol. 7, pp. 17–61.

Franz, S., Leone, M., Ricci-Tersenghi, F., & Zecchina, R. (2001). Exact solutions for diluted spin glasses and optimization problems. *Physical Review Letters*, *87*, 127209–127212.

Gent, I., & Walsh, T. (1999). Beyond NP: the QSAT phase transition. In *Proceedings of the 16th National Conference on Artificial Intelligence (AAAI'99)*, pp. 648–653.

Janson, S. (1987). Poisson convergence and Poisson processes with applications to random graphs. *Stochastic Processes and Applications*, *26*(1), 1–30.

Janson, S., Luczak, T., & Rucinski, A. (2000). *Random graphs*. John Wiley and sons.

Takács, L. (1988). On the limit distribution of the number of cycles in a random graph. *Journal of Applied Probability*, *25*, 359–376.